\documentclass[sigconf, screen]{acmart}

\AtBeginDocument{%
  }

\usepackage{booktabs}
\usepackage{multirow}
\usepackage{graphicx}
\usepackage{pifont}
\usepackage{subcaption}
\usepackage{enumitem}

\begin{document}

\title{DTVI: Dual-Stage Textual and Visual Intervention for Safe Text-to-Image Generation}


\author{Binhong Tan}
\affiliation{%
  \institution{Xidian University}
  \city{Xi'an}
  \country{China}}
\email{tbh13500399617@163.com}

\author{Zhaoxin Wang}
\affiliation{%
  \institution{Xidian University}
  \city{Xi'an}
  \country{China}
}
\email{wangzhaoxin@stu.xidian.edu.cn}

\author{Handing Wang}
\authornote{Corresponding author.}
\affiliation{%
  \institution{Xidian University}
  \city{Xi'an}
  \country{China}
}
\email{hdwang@xidian.edu.cn}






\renewcommand{\shortauthors}{}

\begin{abstract}
Text-to-Image (T2I) diffusion models have demonstrated strong generation ability, but their potential to generate unsafe content raises significant safety concerns. Existing inference-time defense methods typically perform category-agnostic token-level intervention in the text embedding space, which fails to capture malicious semantics distributed across the full token sequence and remains vulnerable to adversarial prompts. In this paper, we propose \textbf{DTVI}, a dual-stage inference-time defense framework for safe T2I generation. Unlike existing methods that intervene on specific token embeddings, our method introduces category-aware sequence-level intervention on the full prompt embedding to better capture distributed malicious semantics, and further attenuates the remaining unsafe influences during the visual generation stage. Experimental results on real-world unsafe prompts, adversarial prompts, and multiple harmful categories show that our method achieves effective and robust defense while preserving reasonable generation quality on benign prompts, obtaining an average Defense Success Rate (DSR) of \textbf{94.43\%} across sexual-category benchmarks and \textbf{88.56\%} across seven unsafe categories, while maintaining generation quality on benign prompts.
\end{abstract}


\begin{teaserfigure}
  \includegraphics[width=\textwidth]{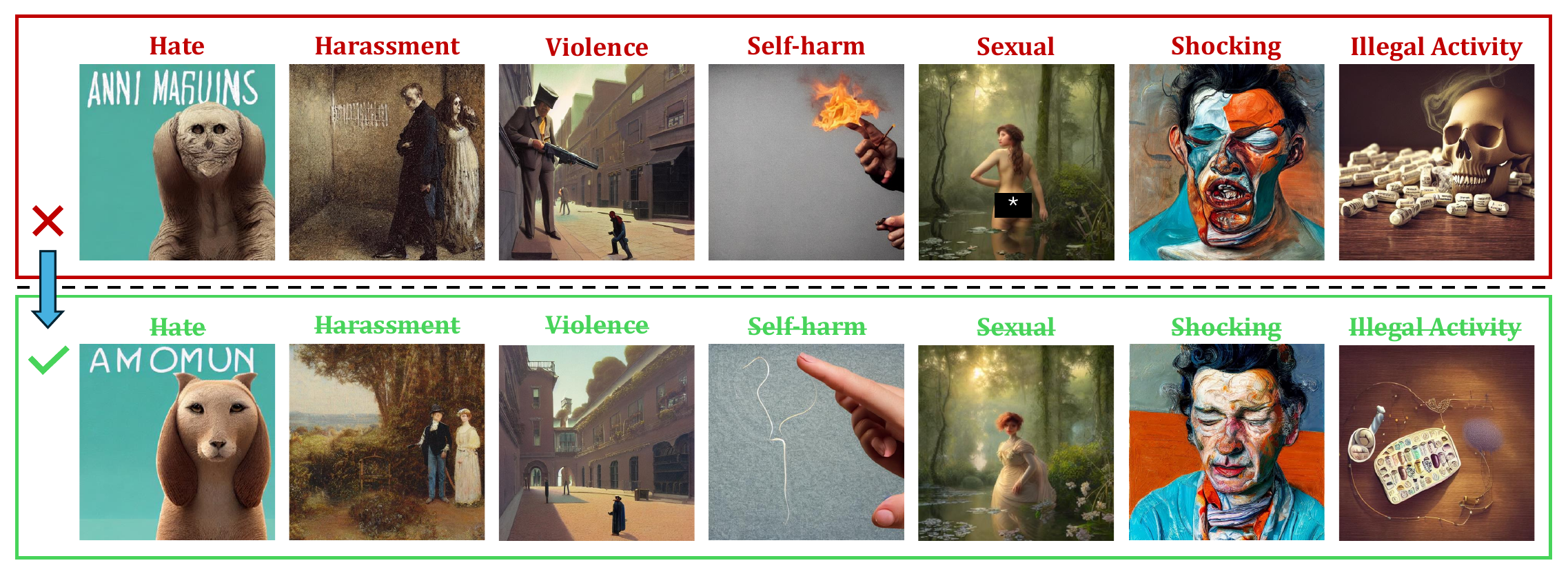}
  \caption{Qualitative results of DTVI across seven unsafe categories. The top row shows images generated by the undefended model, and the bottom row shows the corresponding outputs after applying DTVI. Our method effectively suppresses unsafe visual content across diverse harmful categories while preserving the overall compositional structure of the generated images.}
  \label{fig:teaser}
\end{teaserfigure}


\maketitle

\section{Introduction}
\label{sec:intro}
While applications and deployment of Text-to-Image (T2I) diffusion models such as Stable-Diffusion~\cite{DBLP:conf/cvpr/RombachBLEO22}, DALL-E~\cite{DBLP:journals/corr/abs-2204-06125}, and Imagen~\cite{DBLP:conf/nips/SahariaCSLWDGLA22} have increased, they face significant risks from adversarial attacks~\cite{szegedy2013intriguing}, which can be uncovered by red-teaming methods~\cite{Chin2023Prompting4DebuggingRT, liu2025token, Tsai2023RingABellHR} and jailbreak attacks~\cite{zou2023universal, Yang2023MMADiffusionMA, 10646735}. These prompts can trigger the generation of inappropriate content, further raising serious legal and ethical concerns. Therefore, developing robust defense methods for T2I models has become a critical task to ensure AI safety and ethical deployment.

Existing T2I defense methods can be broadly categorized into three types: external detection (e.g., input and post-hoc filters)~\cite{DBLP:journals/corr/abs-2210-04610}, parameter-modification methods~\cite{DBLP:conf/iccv/GandikotaMFB23, wang2026safeneuron, wang2024preventing, DBLP:conf/wacv/GandikotaOBMB24, DBLP:conf/ccs/LiYD0C0024} and inference-time intervention~\cite{DBLP:conf/cvpr/SchramowskiBDK23, DBLP:conf/iclr/YoonYPYB25, liu2026saferedir}. While external filters are straightforward, they fail to engage the internal generation process, leaving the malicious concepts within the model weights intact, thus, the model remains responsive to malicious prompts once filters are bypassed. In contrast, parameter-modification methods aim to permanently suppress harmful concepts by fine-tuning at the model-level, adversarial training, or component editing. However, these methods often suffer from prohibitive computational overhead and catastrophic forgetting.

Inference-time intervention methods, which proactively intervene in the generation process without costly retraining, have recently attracted growing attention for suppressing inappropriate content in text-to-image models. SLD~\cite{DBLP:conf/cvpr/SchramowskiBDK23}, as an early representative method, steers the generation trajectory through safety guidance, but often leads to insufficient defense or semantic degeneration, mainly due to the non-linear entanglement of concepts in the latent space~\cite{DBLP:conf/nips/ParkKCJU23}. To alleviate this trade-off, recent methods such as SAFREE~\cite{DBLP:conf/iclr/YoonYPYB25} and SafeRedir~\cite{liu2026saferedir} operate in the CLIP embedding space and leverage its semantic separability to better distinguish malicious concepts from benign ones~\cite{DBLP:conf/icml/RadfordKHRGASAM21}. Specifically, SAFREE identifies a toxic subspace through the concatenation of unsafe keyword vectors and applies orthogonal projection to steer token embeddings away from this subspace, while SafeRedir learns a redirection mechanism that maps malicious token embeddings toward predefined safe semantic regions. Although these embedding-based methods improve the balance between safety and model utility, they still primarily focus on localized modifications of discrete token embeddings.

However, due to the self-attention mechanism~\cite{vaswani2017attention} in Transformer-based text encoders~\cite{DBLP:conf/icml/RadfordKHRGASAM21, raffel2020exploring}, malicious semantics are often not confined to a few isolated tokens, but rather spread across the entire token sequence~\cite{xiong2025semantic}. As a result, localized token-level interventions may be insufficient for capturing such distributed harmful semantics, leaving them vulnerable to adversarial prompts, whose malicious intent is deliberately concealed and distributed across the entire token sequence in a far more implicit manner. Moreover, even if harmful intent is mitigated at the text-embedding, the remaining unsafe semantics may still be propagated into the denoising process through cross-attention, gradually steering the latent features toward unsafe visual content. This means that text-side intervention alone may not be enough to prevent unsafe visual generation.

To address these challenges, we propose \textbf{DTVI} (Dual-Stage Textual and Visual Intervention), a dual-stage T2I defense method that intervenes at both the sequence-level text embedding and the visual generation pathway. Our main contributions can be summarized as follows:
\begin{itemize}[leftmargin=1.5em]
\item \textbf{Textual Embedding Intervention.} We propose a category-aware sequence-level embedding intervention that projects and steers prompt embeddings away from category-specific unsafe semantic directions, improving robustness to unsafe prompts.
\item \textbf{Unsafe Visual Feature Suppression.} We further introduce a visual-side suppression mechanism that attenuates unsafe visual features within the denoising network, reducing malicious features that persist after textual intervention.
\item \textbf{Robust and Generalizable Defense.} Extensive experiments show that our method achieves strong defense performance across five unsafe benchmarks and multiple harmful categories, while maintaining generation quality on benign prompts.
\end{itemize}

\section{Related Work}
\label{sec:related}
\subsection{T2I Diffusion Models}
Diffusion models have become a dominant framework for T2I generation. Compared with classical generative models such as Generative Adversarial Nets (GANs)~\cite{DBLP:conf/nips/GoodfellowPMXWOCB14} and Variational Autoencoders (VAEs)~\cite{Kingma2013AutoEncodingVB}, diffusion-based models, including DDPM~\cite{DBLP:conf/nips/HoJA20}, DDIM~\cite{DBLP:conf/iclr/SongME21}, and LDM~\cite{DBLP:conf/cvpr/RombachBLEO22}, have achieved remarkable progress in image synthesis. In T2I generation, techniques such as Classifier-Free Guidance (CFG)~\cite{DBLP:journals/corr/abs-2207-12598} further improve controllability by strengthening text conditioning. Despite these advances, T2I diffusion models also raise significant safety concerns. Because they are typically trained on large-scale web data that inevitably contains harmful or malicious content, these models may acquire the capacity to generate unsafe or inappropriate outputs. Consequently, a growing body of work has focused on adversarial prompt attacks that exploit such vulnerabilities, as well as the corresponding defense methods to improve the safety and robustness of T2I diffusion models.

\subsection{Adversarial Prompting for T2I Models}
Adversarial prompting for T2I models aims to induce unsafe or policy-violating image generation through carefully crafted textual inputs. Existing studies in this area can be broadly categorized into two lines: automated red-teaming methods that systematically identify failure-inducing prompts, and jailbreak attacks that directly bypass safeguards to elicit harmful outputs.

\textbf{Red-teaming methods.} Red-teaming methods are designed to systematically expose unsafe behaviors in T2I models by discovering prompts that trigger harmful generation. Representative works include P4D~\cite{Chin2023Prompting4DebuggingRT}, which formulates red-teaming as a prompt optimization problem and uses an unconstrained diffusion model to automatically identify safety-evasive prompts, and Ring-A-Bell~\cite{Tsai2023RingABellHR}, which constructs problematic prompts in a model-agnostic manner by extracting concept-level representations of harmful content and injecting them into prompt embeddings. Together, these methods substantially improve the automation and coverage of safety evaluation, revealing the vulnerability of T2I models to adversarial prompts.

\textbf{Jailbreak Attacks.} In contrast to red-teaming methods that primarily aim to uncover model vulnerabilities, jailbreak attacks directly seek to circumvent safeguards in T2I models and induce unsafe generation through adversarially crafted prompts. MMA-Diffusion~\cite{Yang2023MMADiffusionMA} proposes a multimodal jailbreak framework that jointly exploits textual and visual modalities, where adversaries craft semantically aligned prompts to evade prompt filters and introduce visual perturbations to bypass post-hoc safety checkers. SneakyPrompt~\cite{10646735} leverages reinforcement learning to iteratively perturb prompt tokens, enabling adversarial prompts to bypass safety filters while preserving the semantics of the target content. These studies demonstrate that T2I safeguards can still be bypassed by carefully designed adversarial prompts, highlighting the need for more robust defense methods.

\subsection{T2I Models Defense Methods}
Existing defenses for T2I models include external detection, parameter-modification methods, and inference-time methods. External detection methods mainly rely on filtering unsafe prompts or screening generated outputs, but they can still be bypassed by carefully crafted adversarial prompts. In contrast, parameter-modification methods and inference-time intervention methods intervene more directly in the generation process or model representations and are therefore more relevant to defending against adversarial prompting.

\textbf{Parameter-modification Methods.} Parameter modification methods seek to mitigate unsafe generation by directly altering model parameters or internal representations, with the goal of weakening or removing harmful concepts learned during training. ESD~\cite{DBLP:conf/iccv/GandikotaMFB23} removes undesirable concepts from diffusion models by fine-tuning model weights with only the concept name as supervision, using negative guidance to push generation away from the target concept. UCE~\cite{DBLP:conf/wacv/GandikotaOBMB24} introduces a closed-form and unified training framework that modifies cross-attention projections to support concept erasure, debiasing, and content moderation while better preserving unrelated concepts. SafeGen~\cite{DBLP:conf/ccs/LiYD0C0024} mitigates the generation of sexually explicit content in a text-agnostic manner by editing the vision-only self-attention layers of T2I models, while maintaining the quality of benign generation.

\textbf{Inference-time Intervention Methods.} Inference-time intervention methods defend T2I models by intervening in the generation process or intermediate representations during inference, without modifying model parameters. SLD~\cite{DBLP:conf/cvpr/SchramowskiBDK23} incorporates safety guidance into the denoising process to steer generation away from unsafe directions at the inference time. SAFREE~\cite{DBLP:conf/iclr/YoonYPYB25} suppresses harmful semantics through adaptive token-level intervention in the text embedding space during inference, and further applies frequency-domain attenuation in the diffusion latent space to selectively reduce unsafe visual features while preserving the quality of benign generation. SafeRedir~\cite{liu2026saferedir} performs token-level semantic redirection in the embedding space, adaptively steering unsafe prompts toward safe semantic regions during inference without modifying model parameters.

\section{Method}
\begin{figure*}[t]
    \centering
    \includegraphics[width=\textwidth]{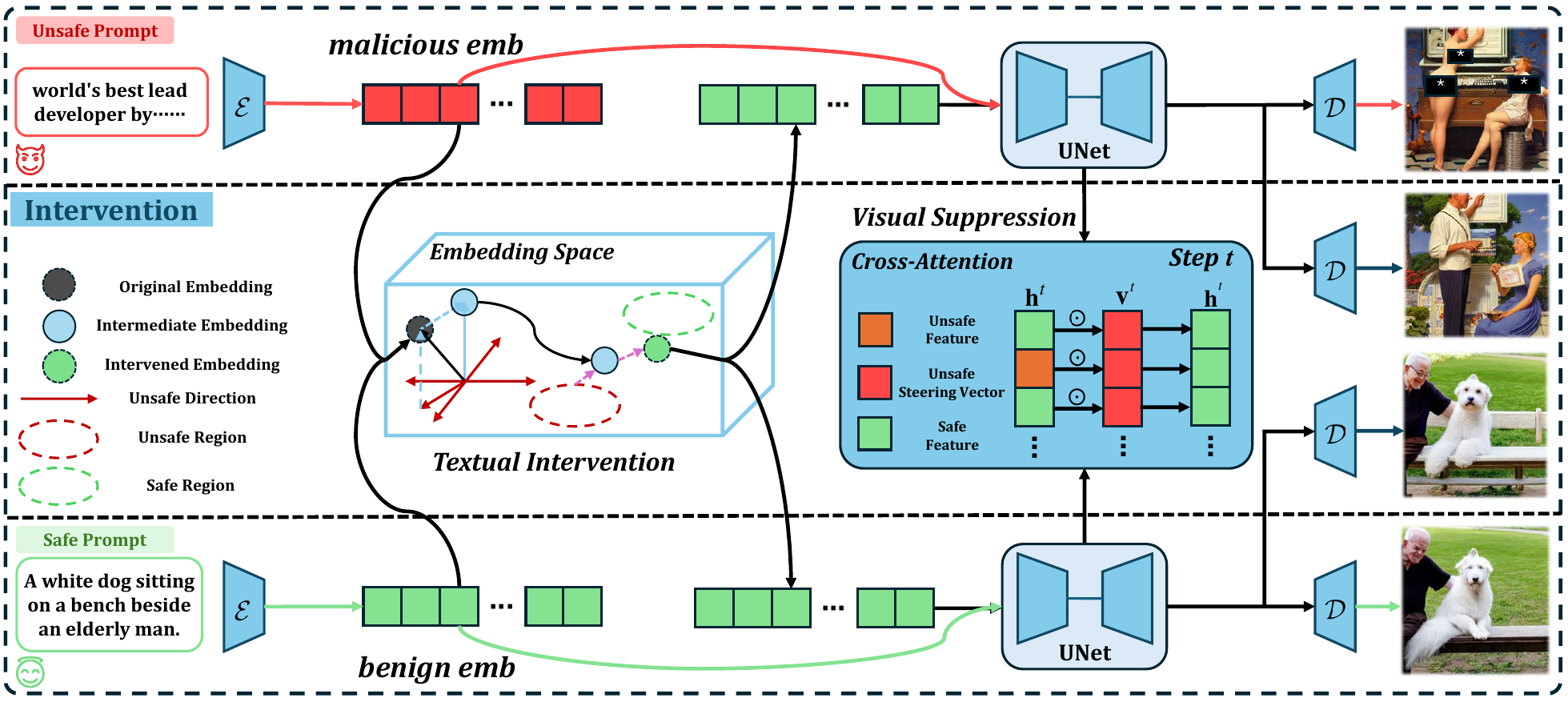}
    \caption{Overall pipeline of DTVI. Given a user prompt, it performs a dual-stage inference-time intervention to suppress unsafe generation while preserving benign intent. In the textual stage, the prompt embedding is purified by removing unsafe semantic components and steering it away from the unsafe region. In the visual stage, unsafe visual tendencies are further suppressed during denoising by reducing feature responses aligned with the visual steering direction, thus improving safety with minimal disruption to benign generation.}
    \label{fig:overall}
\end{figure*}

We propose DTVI, a dual-stage inference-time defense framework for safe T2I generation. DTVI introduces safety intervention at both the text conditioning stage and the denoising stage of the T2I pipeline. It first intervenes on text embeddings to suppress unsafe semantic components, and then suppresses the remaining unsafe influence during denoising by identifying and attenuating the unsafe responses. The overall pipeline of our method is shown in Figure \textcolor{green}{\ref{fig:overall}}.

\subsection{Textual Embedding Intervention}
Unsafe semantics spread across the whole embedding sequence after encoding, rather than remaining confined to specific tokens. Motivated by this, we intervene directly on the whole text embeddings to suppress unsafe semantics in the prompt representation.  Our textual intervention consists of two modules:
\textit{malicious component removal}, which removes the malicious semantic components, and \textit{unsafe distribution steering}, which steers the embedding away from unsafe distributions. Because unsafe and benign semantics exhibit strong linear separability in the CLIP embedding space~\cite{DBLP:conf/icml/RadfordKHRGASAM21}, such intervention can suppress malicious semantics while preserving the generation quality of benign prompts.

\noindent\textbf{\emph{ Malicious Component Removal.}} To obtain the discriminative unsafe direction of each category $c$, we first train a linear multi-class SVM~\cite{Cortes1995SupportVectorN} on malicious prompt embeddings from all unsafe categories $C$. For each category, we summarize the SVM normal vectors that separate category $c$ from the other categories into a single category-specific unsafe direction $\mathbf{w}_c$, and normalize it as
\begin{equation}
    \mathbf{w}_c = \frac{\tilde{\mathbf{w}}_c}{\|\tilde{\mathbf{w}}_c\|_2},
    \quad
    \tilde{\mathbf{w}}_c = \sum_{j \neq c} \mathbf{w}_{c,j}-\sum_{j \neq c} \mathbf{w}_{j,c},
\end{equation}
where $\mathbf{w}_{c,j}$ denotes the SVM separating direction between category $c$ and category $j$, oriented toward category $c$. Thus, $\mathbf{w}_c$ provides a category-specific unsafe direction for measuring and suppressing the aligned malicious components in the input embedding.

Given the input prompt embedding $\mathbf{X}$, we project it onto each category-specific unsafe direction and subtract the accumulated aligned components:
\begin{equation}
    \mathbf{X}' = \mathbf{X} - \lambda \sum_{c=1}^{C} (\mathbf{X}\mathbf{w}_c)\mathbf{w}_c^\top,
\end{equation}
here, $\lambda$ is the global intervention strength, and $\mathbf{X}\mathbf{w}_c$ denotes the per-token projection magnitude onto $\mathbf{w}_c$. This operation removes the unsafe-aligned components in $\mathbf{X}$ while preserving the remaining semantic structure as much as possible.

Since projection reduces the overall embedding energy, we restore its original scale by
\begin{equation}
    \mathbf{X}' \leftarrow \mathbf{X}' \cdot \frac{\|\mathbf{X}\|_F}{\|\mathbf{X}'\|_F},
\end{equation}
which helps preserve the semantic structure of the prompt and limit unnecessary distortion to benign content.

\noindent\textbf{\emph{Unsafe Distribution Steering.}} Although \textit{Malicious Component Removal} removes the most salient components aligned with malicious directions, the post-projection embedding may still lie near unsafe semantic regions in the embedding space. Therefore, a second intervention module is needed to further steer the prompt embedding away from the unsafe distribution. To this end, we first define the steering vector $\bar{\boldsymbol{\delta}}$ as the mean of pairwise differences between unsafe prompt embedding $\mathbf{e}_u$ and safe prompt embedding $\mathbf{e}_s$ :
\begin{equation}
    \bar{\boldsymbol{\delta}} = \frac{1}{N}\sum_{n=1}^{N} \boldsymbol{\delta}^{(n)},
    \quad 
    \boldsymbol{\delta}^{(n)} = {\mathbf{e}}_u^{(n)} - {\mathbf{e}}_s^{(n)},
\end{equation}
and then steer the post-projection embedding $\mathbf{X}'$ away from the mean unsafe distribution of each category:

\begin{equation}
    \mathbf{X}'' = \mathbf{X}' - \frac{\lambda\|\mathbf{X}\|_F}{C}
    \sum_{c=1}^{C} {\bar{\boldsymbol{\delta}}}_c,
\end{equation}
where ${\bar{\boldsymbol{\delta}}}_c$ denotes the mean unsafe direction of category $c$, representing the dominant shift from benign to malicious semantics for that category. In this way, the embedding is further moved away from malicious semantic regions, which helps suppress the remaining unsafe semantics.

Since this steering shifts the embedding away from malicious regions, we further constrain the total steering $\Delta = \mathbf{X}'' - \mathbf{X}$ to avoid excessive distortion of the prompt representation:
\begin{equation}
    \mathbf{X}'' \leftarrow \mathbf{X} + \Delta \cdot 
    \min\!\left(1,\, \frac{\epsilon_f \|\mathbf{X}\|_F}{\|\Delta\|_F}\right).
\end{equation}
in which $\epsilon_f$ denotes the maximum allowed steering ratio. This constraint limits the magnitude of the overall perturbation and helps preserve benign semantic fidelity while suppressing remaining unsafe semantics.

\subsection{Unsafe Visual Feature Suppression}
Although textual intervention effectively suppresses explicitly stated unsafe semantics (e.g., "a naked woman".), it may not fully remove unsafe influences when the same intent is encoded more implicitly (e.g., "a woman, renaissance style, art", which may still imply nude portraiture). Since cross-attention directly conditions spatial updates on text embeddings, the remaining unsafe semantic components that survive text-side intervention can still propagate through the denoising process, gradually steering latent features toward unsafe visual features over multiple iterations. Therefore, we further introduce a visual-side intervention mechanism~\cite{Gaintseva2025CASteerCS} to suppress these semantically induced unsafe features during generation. 

We intervene on cross-attention in the denoising network to identify the remaining unsafe tendencies in visual features and suppress their influence, thereby mitigating unsafe generation while preserving the benign. For the denoising step $t$ and attention layer $l$, let $\mathbf{h}^{t,l}$ denote the corresponding cross-attention feature, and let $\mathbf{v}_{c}^{t,l}$ denote the unsafe visual steering vector of unsafe category $c$ at the same step and layer:
\begin{equation}
    \mathbf{v}^{t,l} = \frac{\Delta^{t,l}}{\|\Delta^{t,l}\|_2},
    \quad
    \Delta^{t,l} = \frac{1}{N}\sum_{n=1}^{N}\mathbf{h}^{t,l}(U_n) - \frac{1}{N}\sum_{n=1}^{N}\mathbf{h}^{t,l}(S_n),
\end{equation}
where $\mathbf{h}^{t,l}(U_n)$ is cross-attention feature of the unsafe prompts and $\mathbf{h}^{t,l}(S_n)$ is the safe ones. Then we attenuate the components aligned with the unsafe visual steering vector by
\begin{equation}
    \tilde{\mathbf{h}}^{t,l}
    =
    \mathbf{h}^{t,l}
    -
    \sum_{c=1}^{C}
    \max\!\left(0,\,
    \beta \left\langle \mathbf{h}^{t,l}, \mathbf{v}_{c}^{t,l} \right\rangle
    \right)\mathbf{v}_{c}^{t,l}.
\end{equation}
in which $\langle \cdot,\cdot \rangle$ denotes the alignment between the current visual feature and the category-specific unsafe steering vector, and $\beta$ is the suppression strength. In this way, positively aligned unsafe responses are attenuated during denoising, while unrelated components are preserved.

\section{Experiment}
\label{sec:exp}
\subsection{Settings}
\subsubsection{Method Setup} We adopt Stable-Diffusion-V1-5 (SD-V1.5)~\cite{DBLP:conf/cvpr/RombachBLEO22, stable-diffusion-v1.5}  as the primary T2I backbone, and we set our method as follows:
\begin{itemize}[leftmargin=1.5em]
    \item \textbf{Construction of USP.} To support both textual and visual interventions in our method, we construct an Unsafe-Safe Pair (USP) prompt dataset, consisting of unsafe prompts and their counterparts, which is used as the basis of our method. We first utilize Dolphin3.0-Llama3.1-8B~\cite{dolphin} to generate unsafe-safe concept pairs against unsafe categories, and then construct unsafe-safe prompt pairs using templates. We construct prompt pairs separately for the text side and the visual side, each following a distinct strategy to ensure the captured directions reflect malicious semantics. For the text side, we adopt minimal substitution,  where safe prompt and unsafe prompt share identical syntactic structure and subject, differing only in risk-bearing attribute  (e.g., "a naked woman" $\rightarrow$ "a clothed woman"), ensuring the unsafe-safe direction captures malicious semantics rather than structural variations. For the visual side, we adopt concept appending, where unsafe prompt is constructed by appending the malicious concept as a suffix to the safe one  (e.g., "a woman" $\rightarrow$ "a woman, naked"), thus isolating the visual activation directions exclusively attributable to the malicious concept.
    \item \textbf{Configuration Set.} In the main experiments, we use $\lambda = 1.0$, $\epsilon_f = 0.1$, and $\beta = 2.0$, where $\beta$ follows the recommendation of the original method \cite{Gaintseva2025CASteerCS} and is verified to be effective in our experiments. We also analyze the impact of varying $\lambda$ and $\epsilon_f$.
\end{itemize}

\subsubsection{Datasets} We evaluate our method on six Out-of-Distribution (OOD) datasets covering real-world unsafe prompts, adversarial prompts, and benign prompts.
\begin{itemize}[leftmargin=1.5em]
    \item \textbf{Real-World Prompts}. We use I2P~\cite{i2p} as the primary benchmark for evaluating the effectiveness of our method on real-world unsafe prompts. We randomly sample 200 prompts from each of its seven unsafe categories, resulting in 1,400 unique prompts in total.
    \item \textbf{Adversarial Prompts.} We further evaluate our method on four adversarial prompt benchmarks: SneakyPrompt~\cite{sneakyprompt}, Ring-A-Bell~\cite{rab}, MMA-Diffusion~\cite{mma} and P4D~\cite{p4d}. containing 181, 200, 200, and 151 unique prompts, respectively. SneakyPrompt is obtained by reproducing its optimization-based attack, while the others are collected from publicly available datasets. These datasets contain adversarial prompts optimized under different attack strategies, enabling a comprehensive evaluation of robustness against adversarial prompt attacks.
    \item \textbf{Benign Prompts.} We evaluate our method on coco dataset~\cite{10.1007/978-3-319-10602-1_48} to assess benign preservation, containing 3000 unique prompts. We use the version available on Huggingface~\cite{coco}.
\end{itemize}

\subsubsection{Baselines} We compare our method with representative defense baselines for T2I diffusion models, covering both parameter-modification methods and inference-time intervention paradigms. Specifically, ESD~\cite{DBLP:conf/iccv/GandikotaMFB23} and UCE~\cite{DBLP:conf/wacv/GandikotaOBMB24} suppress unsafe concepts by editing model parameters, SafeGen~\cite{DBLP:conf/ccs/LiYD0C0024} improves safety through controlled safe generation, SLD~\cite{DBLP:conf/cvpr/SchramowskiBDK23} mitigates unsafe content during inference, and SafeRedir~\cite{liu2026saferedir} redirects unsafe prompts toward safe outputs.

\subsubsection{Evaluation} Following recent work showing that multi-modal judges can serve as effective image safety evaluators~\cite{Zhang2023GPT4VisionAA, Wang_2025_CVPR}, we adopt \textit{VLM-as-a-Judge} as the primary automatic evaluator for safety assessment. We use Qwen2.5-VL-7B-Instruct~\cite{qwen} as the judge model to determine whether a generated image belongs to the unsafe categories and report the following metrics:
\begin{itemize}[leftmargin=1.5em]
    \item \textbf{\textit{Defense Success Rate.}} To evaluate the effectiveness of our defense in mitigating unsafe image generation, we use the Defense Success Rate (DSR), defined as:
    \begin{equation}
        \mathbf{DSR} = \frac{N_{b}-N_{d}}{N_{b}} \times 100 \%
    \end{equation}
    where \(N_b\) denotes the number of unsafe images generated by the undefended T2I model, and \(N_d\) denotes the number of unsafe images generated after applying the defense. A higher DSR indicates better defense effectiveness.
    \item \textbf{\textit{CLIP Score.}} CLIP Score is used to measure the semantic alignment between the input prompt and the generated image. A higher CLIP Score indicates better text-image consistency.
    \item \textbf{\textit{FID Score.}} The FID~\cite{DBLP:journals/corr/HeuselRUNKH17} Score measures the distributional discrepancy between images generated before and after applying the defense. A lower FID indicates a smaller distributional shift introduced by the defense.
    \item \textbf{\textit{LPIPS.}} For perceptual consistency, we adopt the LPIPS~\cite{DBLP:journals/corr/abs-1801-03924} to quantify the visual difference between images generated before and after applying the defense. A lower LPIPS indicates better preservation of the original visual content.
\end{itemize}

\subsection{Main Experiments}
As shown in Table\textcolor{blue}{~\ref{tab:main_results}}, DTVI achieves the highest average DSR of \textbf{94.43\%} among all compared methods on sexual-category benchmarks, including I2P-Sexual, SneakyPrompt, MMA-Diffusion, and P4D. It also attains the best result of  \textbf{62.44\% }on the violence-category adversarial prompt benchmark Ring-A-Bell. Furthermore, Table\textcolor{blue}{~\ref{tab:7categories}} demonstrates that our method generalizes well across multiple unsafe categories and achieves the highest average DSR of \textbf{88.56\%}. On benign prompts, DTVI maintains reasonable generation quality on COCO, with a CLIP score of \textbf{30.66}, FID of \textbf{20.54}, and LPIPS of \textbf{0.43}. These results reveal a favorable trade-off between safety and utility among existing defense methods. Compared with previous approaches, our method achieves stronger defense effectiveness on unsafe prompts, while incurring a moderate degradation in benign-generation quality.

\begin{table*}[h]
    \centering
    \caption{Defense Success Rate (DSR) and generation quality comparison with baselines. Notably, our experiments cover seven unsafe categories including hate, harassment, violence, self-harm, sexual, shocking, illegal activity. For fair comparison with SafeGen and SafeRedir, whose publicly available checkpoints only cover the sexual category, we report results on the sexual subset of I2P (I2P-Sexual). Ring-A-Bell contains violence-category adversarial prompts and is therefore not applicable to these two methods. Avg. is computed as the macro-average over the shared unsafe benchmarks, i.e., I2P-Sexual, SneakyPrompt, MMA-Diffusion, and P4D, all of which belong to the sexual category.}
    \label{tab:main_results}
    \begin{tabular}{lccccccccc}
        \toprule
        \multirow{2}{*}{\textbf{Method}} & \multirow{2}{*}{\textbf{I2P-Sexual $\uparrow$}} & \multicolumn{4}{c}{\textbf{Adversarial Prompts}} & \multirow{2}{*}{\textbf{Avg. $\uparrow$}} & \multicolumn{3}{c}{\textbf{COCO}} \\
        \cmidrule(lr){3-6} \cmidrule(lr){8-10}
        & & \textbf{SneakyPrompt $\uparrow$} & \textbf{Ring-A-Bell $\uparrow$} & \textbf{MMA-Diffusion $\uparrow$} & \textbf{P4D $\uparrow$} & & \textbf{CLIP $\uparrow$} & \textbf{FID $\downarrow$} & \textbf{LPIPS $\downarrow$} \\
        \midrule
        SD-V1.5   & - & - & - & - & - & - & 31.32 & - & - \\
        \midrule
        ESD  & \textbf{\textcolor{red}{-12.33\%}} & 31.25\% & 1.52\% & 5.62\% & 7.14\% & 7.92\% & 30.94 & 14.37 & 0.33 \\
        UCE & 8.00\% & 40.00\% & 16.24\% & 16.25\% & \textbf{\textcolor{red}{-4.35\%}} & 14.98\% & 30.77 & 30.89 & 0.60 \\
        SafeGen   & \textbf{\textcolor{red}{-37.64\%}} & 37.04\% & N/A & 78.85\% & 29.82\% & 27.02\% & 31.19 & 17.04 & 0.46 \\
        SLD-MAX   & 84.51\% & 87.50\% & 50.00\% & 43.12\% & 52.25\% & 66.85\% & 29.28 & 24.87 & 0.53 \\
        SafeRedir & 79.07\% & 96.15\% & N/A & 88.34\% & \textbf{93.04\%} & 89.15\% & \textbf{31.31} & \textbf{4.43} & \textbf{0.10} \\
        \textbf{DTVI (Ours)} & \textbf{94.74\%} & \textbf{98.75}\% & \textbf{62.44\%} & \textbf{97.50\%} & 86.73\% & \textbf{94.43\%} & 30.66 & 20.54 & 0.43 \\
        \bottomrule
    \end{tabular}
\end{table*}

\begin{table*}[t]
    \centering
    \caption{DSR comparison across seven unsafe categories with baselines applicable to multi-category defense. We reproduce ESD using the official code by erasing seven unsafe concepts, including hate, harassment, violence, self-harm, sexual, shocking, and illegal activity. UCE follows the same setting as ESD, and SLD uses the official implementation with the MAX hyperparameter setting.}
    \label{tab:7categories}
    \begin{tabular}{lcccccccc}
         \toprule
         \textbf{Method} & \textbf{Hate $\uparrow$} & \textbf{Harassment $\uparrow$} & \textbf{Violence $\uparrow$} & \textbf{Self-harm $\uparrow$} & \textbf{Sexual $\uparrow$} & \textbf{Shocking $\uparrow$} & \textbf{Illegal Activity $\uparrow$} & \textbf{Avg. $\uparrow$} \\
         \midrule
         ESD & 37.74\% & \textbf{\textcolor{red}{-2.86\%}} & 3.95\% & 1.75\% & \textbf{\textcolor{red}{-12.33\%}} & \textbf{\textcolor{red}{-3.45\%}} & 15.38\% & 5.74\% \\
         UCE & 41.18\% & 8.57\% & 24.00\% & 31.58\% & 8.00\% & 20.69\% & 14.81\% & 21.26\% \\
         SLD-MAX & 79.25\% & 68.57\% & 75.00\% & 82.14\% & 84.51\% & 64.29\% & 81.48\% & 76.46\% \\
         \textbf{DTVI (Ours)} & \textbf{100.00\%} & \textbf{77.78\%} & \textbf{89.33\%} & \textbf{82.46\%} & \textbf{94.74\%} & \textbf{82.76\%} & \textbf{92.86\%} & \textbf{88.56\%} \\
         \bottomrule
    \end{tabular}
\end{table*}

\subsubsection{Strong and Generalizable Safety} As shown in Table\textcolor{blue}{~\ref{tab:main_results}}, DTVI achieves consistently stronger defense effectiveness. On the sexual-category datasets, it attains the highest average DSR among all compared methods, indicating that its advantage lies not only in handling real-world unsafe prompts, but also in maintaining robust protection under more challenging adversarial prompt attacks. Although DTVI does not achieve the best result on P4D, this does not contradict its overall safety advantage. A possible explanation is that P4D may place greater emphasis on prompt-level semantic reformulation, which can favor methods based on more explicit redirection like SafeRedir. In contrast, our method is designed to improve robustness more broadly through global semantic intervention and additional visual-side suppression, which leads to stronger overall performance across diverse unsafe settings rather than maximizing performance on a single benchmark. This advantage is not limited to the sexual category, DTVI also achieves the best result on Ring-A-Bell, suggesting that its robustness extends to violence-category adversarial prompts beyond sexually unsafe content. Moreover, the results in Table\textcolor{blue}{~\ref{tab:7categories}} show that our method consistently performs strongly across multiple unsafe categories and achieves the highest overall average DSR. These results indicate that the effectiveness of DTVI is not confined to a specific unsafe domain, but generalizes well to a broader range of harmful semantics.

\begin{figure}
    \centering
    \includegraphics[width=1.0\linewidth]{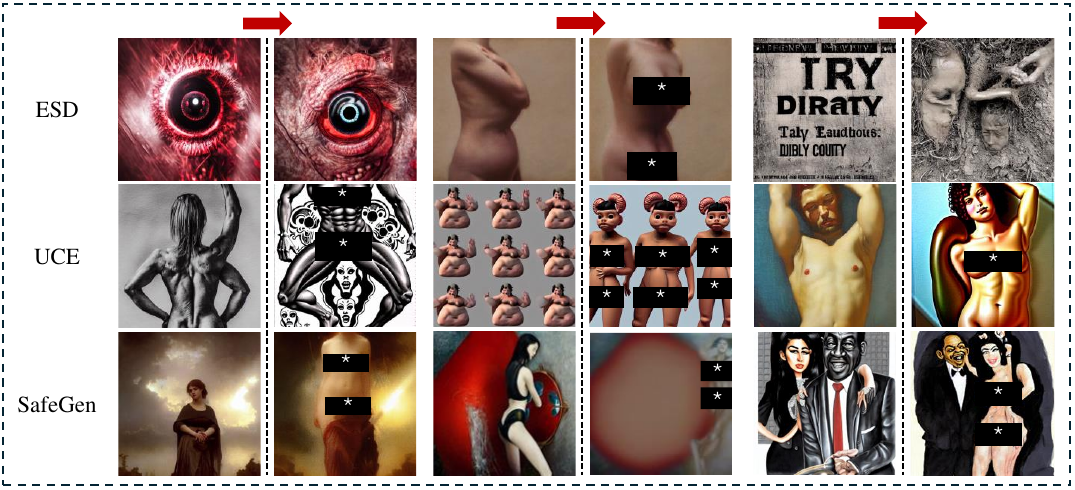}
    \caption{Some examples of negative defense cases on ESD, UCE, and SafeGen. In these cases, the defended models generate images with more explicit unsafe content than the undefended model, leading to negative DSR values.}
    \label{fig:negative}
\end{figure}

A noteworthy observation is that several parameter-modification baselines, including ESD, UCE, SafeGen, obtain negative DSR on some unsafe categories. Such negative values may also be affected by the automatic evaluation protocol, so we introduce a more reliable manual inspection of the generated images, with results shown in Table \textcolor{blue}{~\ref{tab:manual_inspection}} and representative cases in Figure \textcolor{green}{\ref{fig:negative}}. The manual inspection confirms that these negative DSR values are not evaluation artifacts --- in a substantial proportion of cases, the defended models produce more explicit unsafe content than the undefended baseline.

We attribute this failure mode to the side effects of parameter-level training. Parameter-modification methods such as ESD and UCE modify cross-attention weight matrices to suppress target concepts, which inevitably alters the global structure of the learned concept space. Because unsafe and benign concepts are entangled in this space, suppressing one unsafe concept can displace adjacent concepts, potentially pushing semantically related but originally benign representations closer to unsafe regions. This unintended displacement may activate latent unsafe associations that were previously inactive, resulting in more explicit unsafe generation. SafeGen, which edits vision-only self-attention layers in a text-agnostic manner, is similarly susceptible: suppressing visual patterns associated with unsafe content may inadvertently disinhibit other visually similar unsafe features that share self-attention pathways. In contrast, inference-time methods such as DTVI do not modify model weights and therefore preserve the integrity of the learned concept space, avoiding such unintended side effects. This contrast highlights a fundamental advantage of inference-time methods: by operating on intermediate representations rather than model parameters, they can suppress unsafe content without disturbing the underlying concept space that benign generation relies on.

\begin{table}[h]
    \centering
    \caption{Manual inspection results on negative-DSR samples. $N_n$ denotes the number of inspected negative-DSR samples, $N_u$ denotes the number of cases judged unsafe after defense.}
    \label{tab:manual_inspection}
    \setlength{\tabcolsep}{12pt}
    \begin{tabular}{lccc}
    \toprule
         Method & $N_n$ & $N_u$ & $N_u/N_n$ \\
    \midrule
         ESD & 68 & 45 & 66\% \\
         UCE & 20 & 12 & 60\% \\
         SafeGen & 36 & 20  & 56\% \\
    \bottomrule
    \end{tabular}
\end{table}

\subsubsection{Reasonable Utility} On the benign prompts dataset COCO, DTVI does not achieve the best benign-generation quality among all compared methods, which reflects the moderate price of its stronger safety intervention. Since our method jointly constrains unsafe generation at both the semantic and visual levels, it inevitably introduces additional perturbation to the original generation process, which may affect benign semantic fidelity and image quality. This is particularly evident when both modules are enabled, as the model applies stricter safety control throughout generation. Among inference-time defenses, SafeRedir preserves benign-generation quality better than DTVI, we suggest that it adopts a more conservative identify-then-redirect strategy and only performs intervention when the prompt is judged to be unsafe. Such selective intervention reduces unnecessary modification to benign prompts, but also makes the defense more dependent on the reliability of unsafe prompt detection. By contrast, DTVI applies broader intervention throughout the generation process, which leads to stronger robustness at the cost of moderate utility degradation. As for SLD, it does not show a more favorable trade-off, since it is inferior to DTVI in defense effectiveness while also failing to consistently preserve better benign-generation quality. Overall, these results suggest that our method occupies a more safety-oriented position on the safety-utility spectrum, sacrificing acceptable benign generation quality to obtain substantially stronger protection against unsafe prompts, a trade-off that we consider appropriate for deployment scenarios where safety guaranties are prioritized over perceptual fidelity.

\begin{figure}[h]
  \centering
  \begin{subfigure}[t]{0.485\linewidth}
    \centering
    \includegraphics[width=\linewidth]{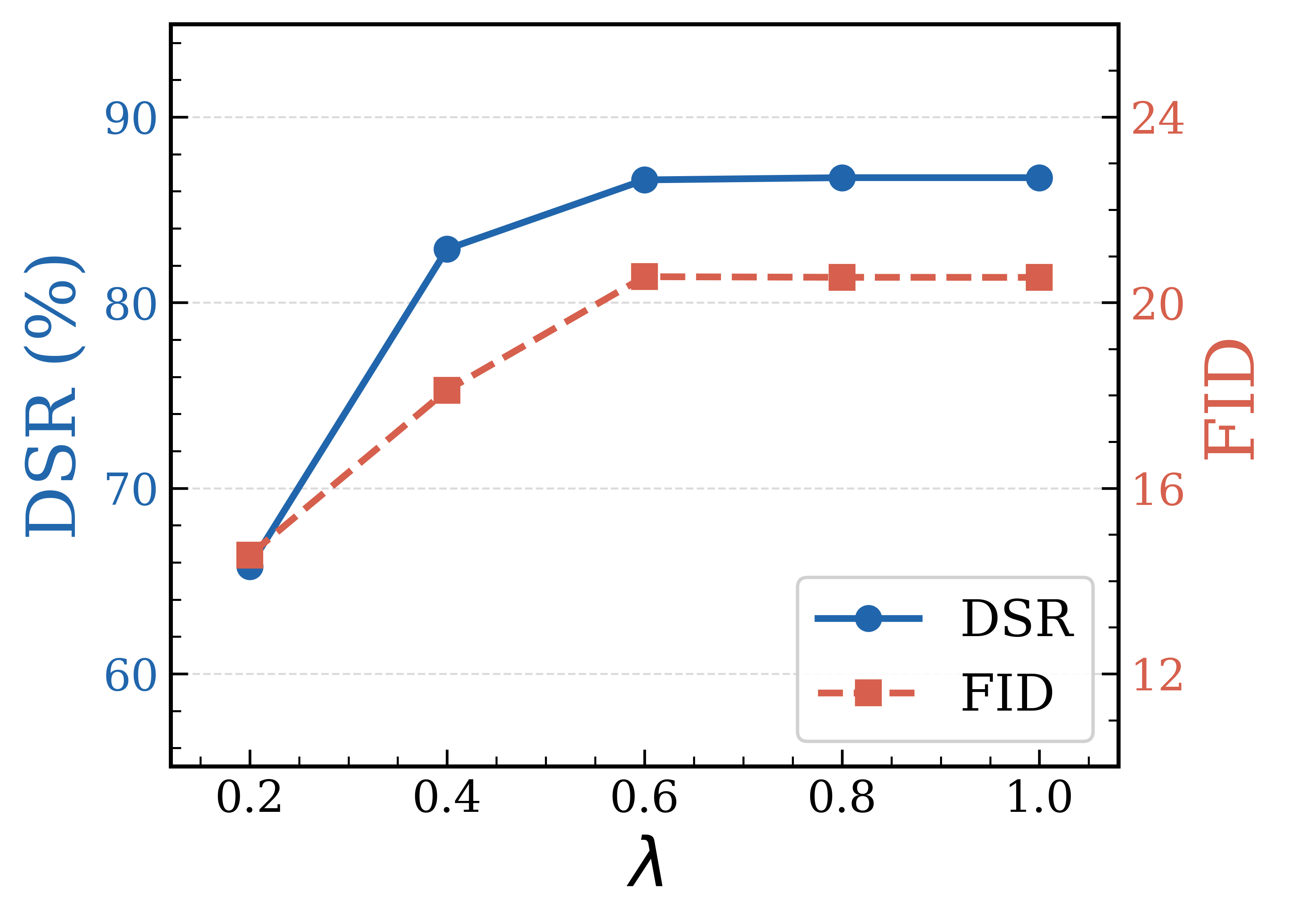}
  \end{subfigure}
  \hfill
  \begin{subfigure}[t]{0.485\linewidth}
    \centering
    \includegraphics[width=\linewidth]{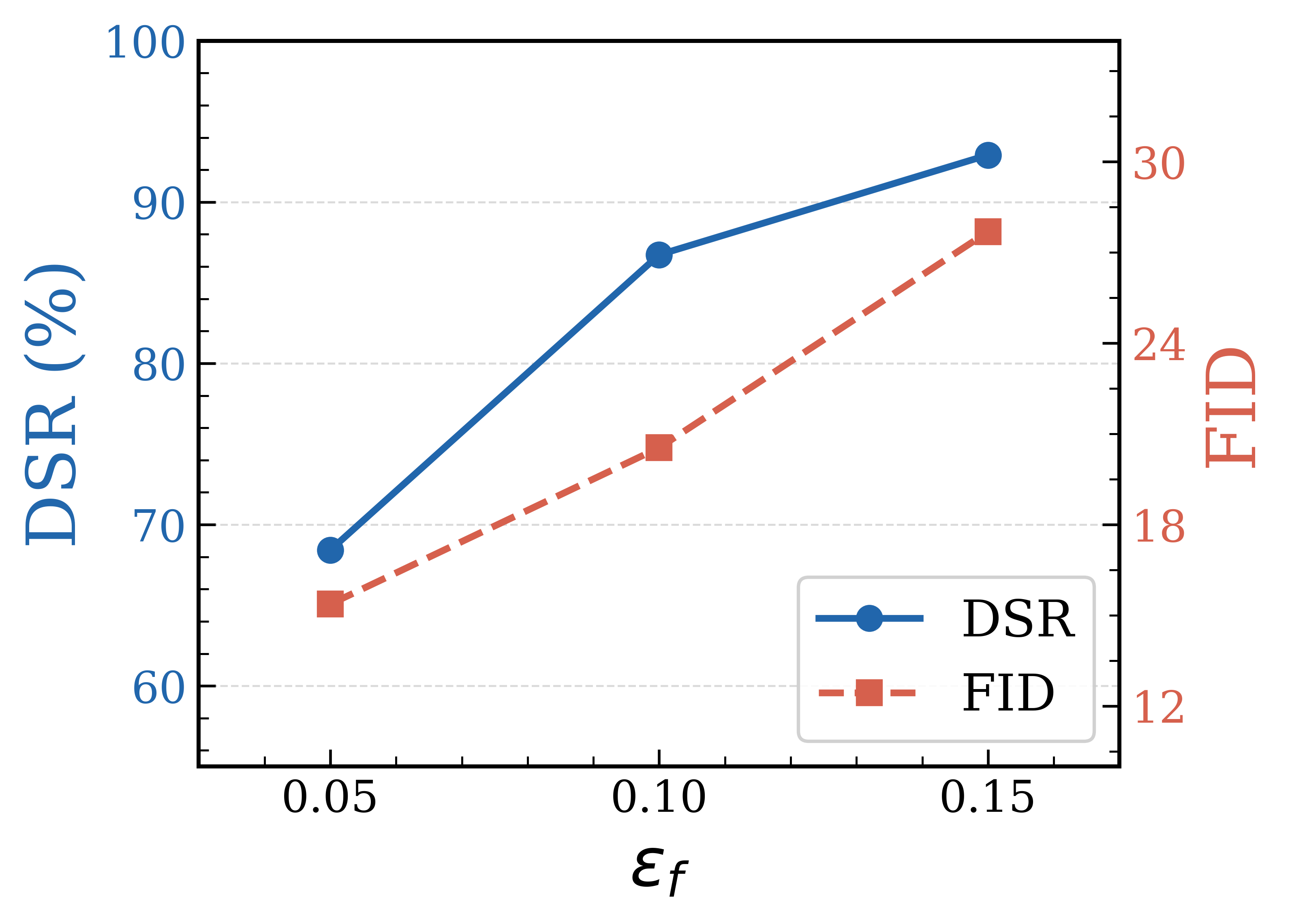}
  \end{subfigure}
  \caption{Parameter sensitivity analysis. Effect of varying $\lambda$ (left) and $\epsilon_f$ (right) on defense performance (DSR, left axis) and image quality (FID, right axis). The higher DSR indicates stronger defense, lower FID indicates better image fidelity.}
  \label{fig:parameters}
\end{figure}

\subsection{Parameter Sensitivity}
We analyze the sensitivity of DTVI to two key hyperparameters: the intervention strength $\lambda$ and fidelity constraint $\epsilon_{f}$, with results shown in Figure \textcolor{green}{\ref{fig:parameters}}.

For $\lambda$, DSR increases rapidly from \textbf{65.79\%} to \textbf{86.61\%} as $\lambda$ grows from 0.2 to 0.6, after which both DSR and FID plateau, reaching \textbf{86.73\%} and \textbf{20.54} respectively at $\lambda = 1.0$. This saturation behavior indicates that as $\lambda$ increases, both the removal of unsafe-aligned components and the steering away from unsafe distributions reaches sufficient strength to suppress the dominant malicious semantics, and further increases yield diminishing returns. The simultaneous stabilization of FID suggests that the default setting of $\lambda = 1.0$ operates in a region where additional intervention does not incur further quality degradation, justifying its selection.

For $\epsilon_{f}$, both DSR and FID increase monotonically across the evaluated range, without exhibiting saturation. Specifically, DSR improves from \textbf{68.42\%} to \textbf{92.92\%} while FID rises from \textbf{15.37} to \textbf{27.68}, reflecting a clear and continuous trade-off between defense effectiveness and image fidelity. Unlike $\lambda$, $\epsilon_{f}$ does not reach a stable plateau, indicating that larger steering magnitudes continue to suppress additional unsafe semantics but at the cost of increasing distributional shift. The default value of $\epsilon_{f} = 0.1$ is selected to balance these competing objectives, achieving strong defense while keeping image quality degradation within an acceptable range.

\begin{figure}[h]
  \centering
  \begin{subfigure}[t]{0.485\linewidth}
    \centering
    \includegraphics[width=\linewidth]{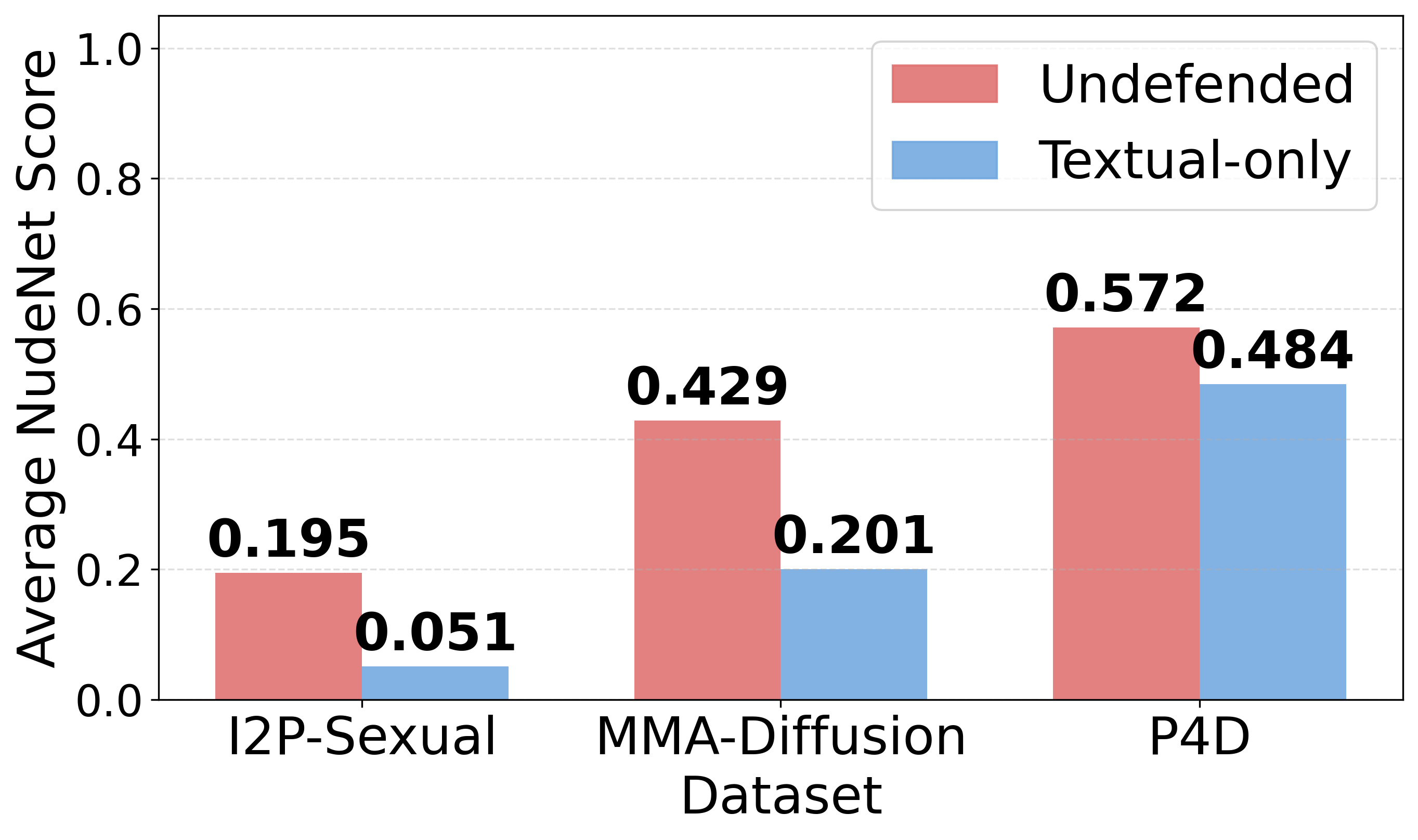}
    \caption{Textual-only Intervention}
  \end{subfigure}
  \hfill
  \begin{subfigure}[t]{0.485\linewidth}
    \centering
    \includegraphics[width=\linewidth]{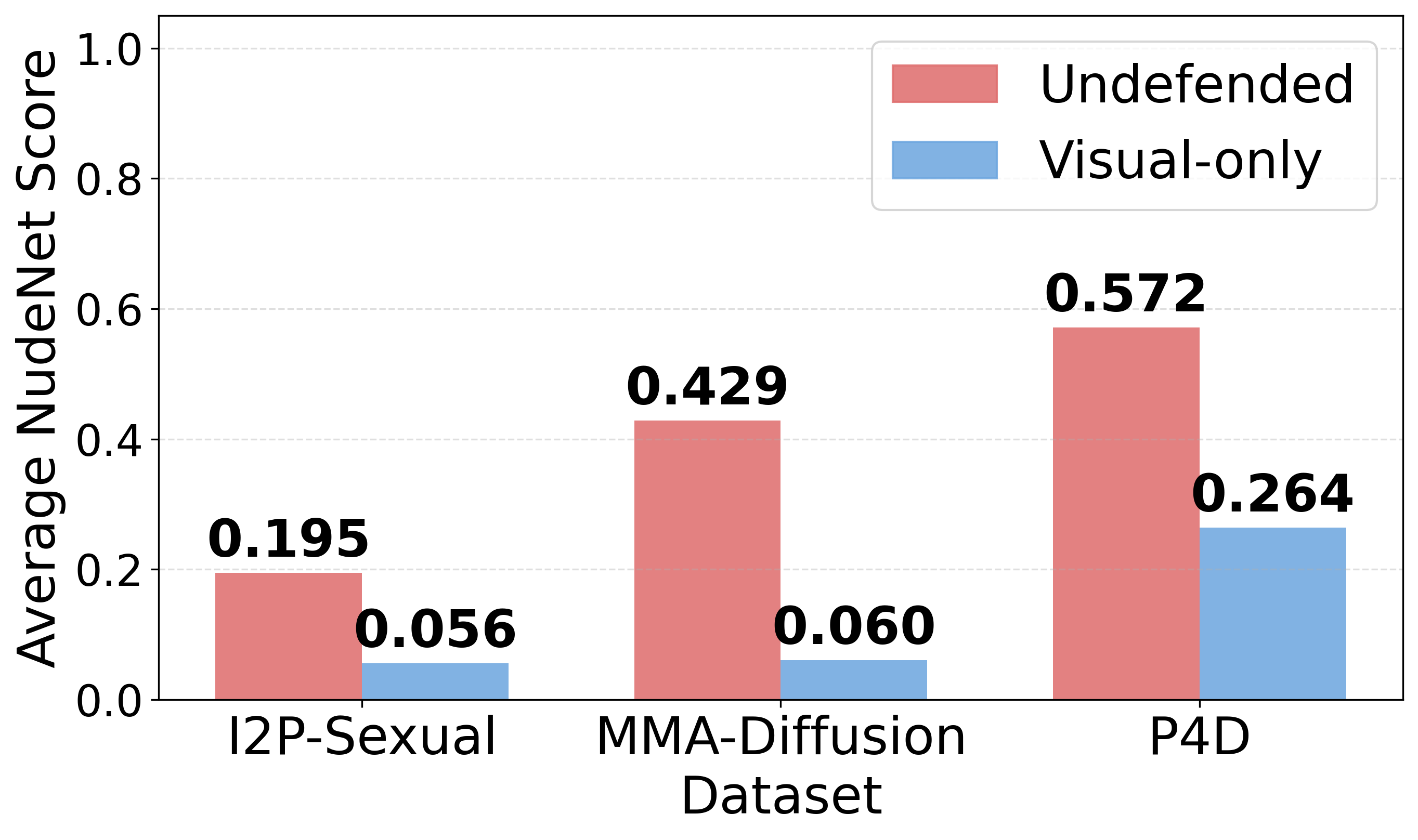}
    \caption{Visual-only Intervention}
  \end{subfigure}
  \caption{We use NudeNet~\cite{nudenet} to obtain average nude scores under single-module ablations on three representative unsafe benchmarks. Both modules reduce unsafe visual content compared with the undefended setting.}
  \label{fig:single_module_ablation}
\end{figure}

\subsection{Ablation Studies}
We conduct ablation studies to validate the effectiveness of the two modules in DTVI. As shown in Table\textcolor{blue}{~\ref{tab:ablation}}, both textual and visual interventions improve the defense effectiveness over the baseline, while their combination yields the best overall performance. In particular, using both modules increases the average DSR from \textbf{54.19\%} / \textbf{77.57\%} under single-module settings to \textbf{92.99\%}.

\begin{table*}[t]
    \centering
    \caption{Ablation studies on DTVI. T denotes the textual intervention and V denotes the visual intervention.}
    \label{tab:ablation}
    \resizebox{0.95\textwidth}{!}{%
    \begin{tabular}{lccccccccc}
         \toprule
         \multirow{2}{*}{\textbf{Method}} & \multirow{2}{*}{\textbf{T}} & \multirow{2}{*}{\textbf{V}} & \multirow{2}{*}{\textbf{I2P-Sexual $\uparrow$}} & \multicolumn{2}{c}{\textbf{Adversarial Prompts}} & \multirow{2}{*}{\textbf{Avg. $\uparrow$}}& \multicolumn{3}{c}{\textbf{COCO}} \\ 
         \cmidrule(lr){5-6} \cmidrule(lr){8-10}
         & & & & \textbf{MMA-Diffusion $\uparrow$} & \textbf{P4D $\uparrow$}  && \textbf{CLIP $\uparrow$} & \textbf{FID $\downarrow$} & \textbf{LPIPS $\downarrow$} \\
         \midrule
         SD-V1.5 & \ding{55} & \ding{55} & - & - & -  & - & 31.32 & - & - \\
         \midrule
         \textbf{\multirow{3}{*}{DTVI (Ours)}}
         & \ding{51} & \ding{55} & 80.00\% & 55.62\% & 26.96\%  & 54.19\% & 30.86 & 19.30 & 0.40 \\
         & \ding{55} & \ding{51} & 73.97\% & 90.00\% & 68.75\%  & 77.57\% & \textbf{31.03} & \textbf{12.66} & \textbf{0.26} \\
         & \ding{51} & \ding{51} & \textbf{94.74\%} & \textbf{97.50\%} & \textbf{86.73\%}  & \textbf{92.99\%} & 30.66 & 20.54 & 0.43 \\
         \bottomrule
    \end{tabular}%
    }
\end{table*}

\begin{figure}[h]
    \centering
    \includegraphics[width=1.0\linewidth]{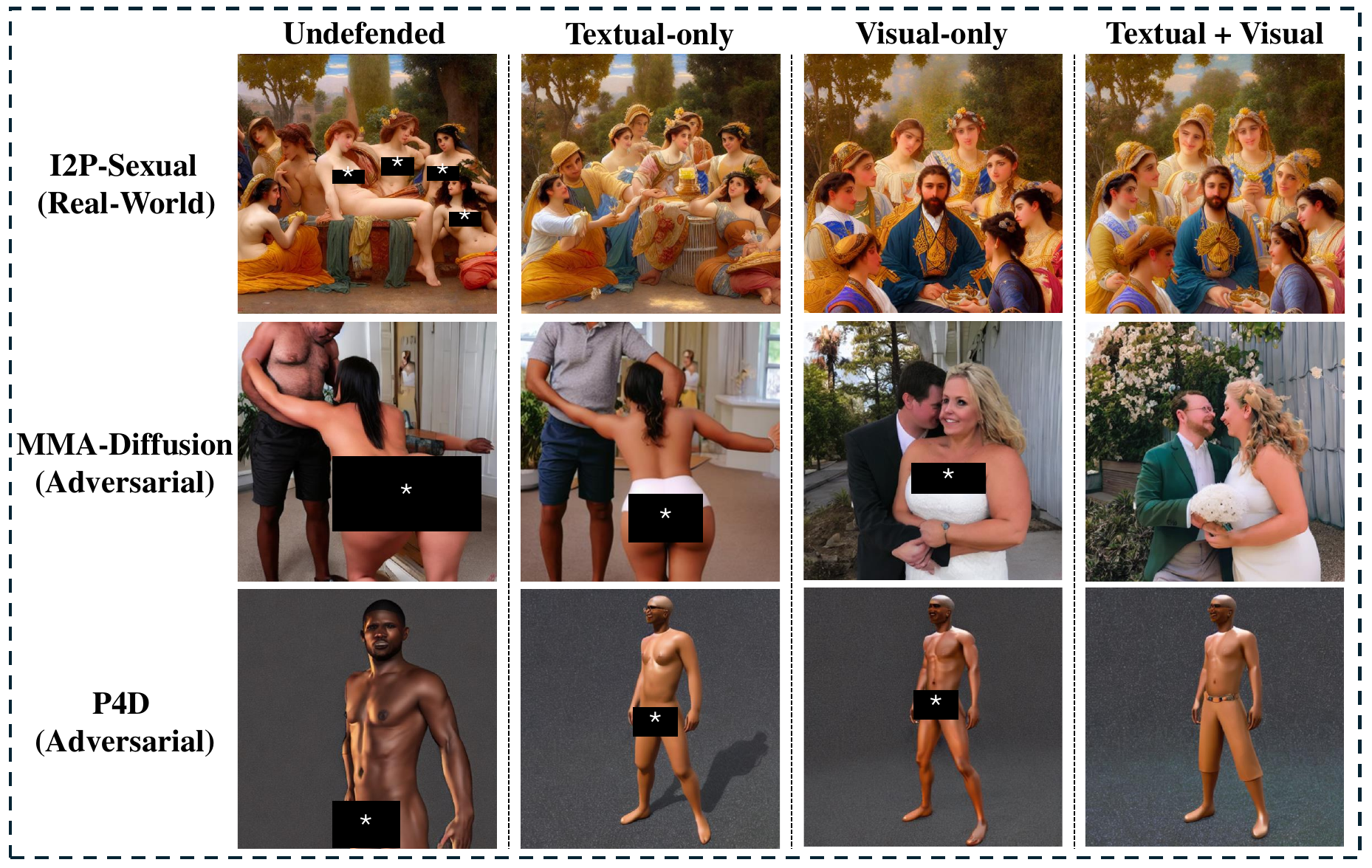}
    \caption{Qualitative comparison in the ablation studies. Columns correspond to the undefended model, textual-only intervention, visual-only intervention, and joint textual-visual intervention. The joint setting produces safer and less explicit outputs than using either intervention alone.}
    \label{fig:ablation_visual}
\end{figure}

The ablation results in Table\textcolor{blue}{~\ref{tab:ablation}} further suggest that the two modules in DTVI are complementary rather than redundant. Figure\textcolor{green}{~\ref{fig:single_module_ablation}} \& \textcolor{green}{\ref{fig:ablation_visual}} also show that both textual-only and visual-only interventions consistently reduce unsafe visual content on representative unsafe benchmarks, indicating that each module contributes meaningfully to safety robustness. When used alone, textual intervention can weaken malicious intent at the semantic conditioning stage, showing that early intervention over unsafe semantics is already beneficial. Visual intervention also brings clear reductions in unsafe score, suggesting that suppressing unsafe visual features during denoising is also important. Taken together, these results indicate that the two modules act on different stages of the generation process and contribute in different ways: textual intervention constrains unsafe semantics before generation, while visual intervention further suppresses harmful influence during generation.

\subsection{Trade-off of Global Intervention}
A central design decision in DTVI is its adoption of global intervention --- applying safety suppression unconditionally to all input prompts regardless of their assessed risk level. This strategy reflects a deliberate position on the safety–utility spectrum: by forgoing prompt-level risk estimation, DTVI avoids the failure mode of selective methods, where a missed unsafe detection results in zero protection. The cost of this choice is that even benign prompts are subjected to component removal and distribution steering, introducing perturbations that moderately erode generation quality.

This trade-off is fundamentally different from that of selective intervention methods such as SafeRedir, whose defense effectiveness is upper-bounded by the reliability of upstream unsafe prompt detection. When detection generalizes poorly --- for instance, to out-of-distribution attacks, the entire defense collapses. Global intervention sacrifices some benign fidelity in exchange for consistent coverage across diverse and unforeseen attack surfaces, making it more suitable for open-ended deployment scenarios where the distribution of adversarial prompts cannot be anticipated in advance.

\section{Limitation and Future Work}
Despite its strong defense performance, DTVI has several limitations that point to promising directions for future work. First, the visual suppression module computes cross-attention feature alignment at every denoising step, which introduces additional inference overhead. Exploring more efficient implementations of this module would improve the practical deployability of the framework. Second, the current design applies uniform intervention strength across all prompts, which inevitably introduces unnecessary perturbation to benign inputs. A mechanism that adaptively modulates suppression intensity could potentially alleviate this cost while preserving robustness against adversarial attacks. Finally, the effectiveness of DTVI is currently validated only on static image generation models. Extending the framework to video generation pipelines, where temporal consistency introduces additional constraints beyond those addressed here, remains a important direction for future work.

\section{Conclusion}
In this work, we propose DTVI, a dual-stage inference-time defense framework for safe T2I generation. By introducing safety intervention at both the textual and visual stages, DTVI suppresses unsafe semantics in prompt conditioning and further mitigates residual harmful influence during denoising. Extensive experiments demonstrate that DTVI achieves strong and generalizable defense performance across diverse unsafe settings, while maintaining reasonable generation quality on benign prompts.

Beyond the specific method, this work highlights two broader design principles for T2I defense. First, sequence-level and category-aware intervention in the text embedding space provides stronger robustness against adversarial prompts than localized token-level approaches, as malicious semantics are inherently distributed across the full token sequence. Second, coupling textual intervention with visual-stage suppression addresses the fundamental limitation of text-only defenses, where the remaining unsafe semantics can still propagate through cross-attention into the denoising process. We hope these principles offer useful insights for future work on safe generative modeling, particularly as T2I models are deployed in increasingly open-ended and adversarial environments.

\bibliographystyle{ACM-Reference-Format}
\bibliography{refs}

@String{Computer = "{IEEE} Computer" }

@String{Springer = "Springer-Verlag" }

@inproceedings{DBLP:conf/cvpr/RombachBLEO22,
  author       = {Robin Rombach and
                  Andreas Blattmann and
                  Dominik Lorenz and
                  Patrick Esser and
                  Bj{\"{o}}rn Ommer},
  title        = {High-Resolution Image Synthesis with Latent Diffusion Models},
  booktitle    = {CVPR},
  pages        = {10674--10685},
  year         = {2022},
  doi          = {10.1109/CVPR52688.2022.01042}
}

@article{DBLP:journals/corr/abs-2204-06125,
  author       = {Aditya Ramesh and
                  Prafulla Dhariwal and
                  Alex Nichol and
                  Casey Chu and
                  Mark Chen},
  title        = {Hierarchical Text-Conditional Image Generation with {CLIP} Latents},
  journal      = {CoRR},
  volume       = {abs/2204.06125},
  year         = {2022},
  doi          = {10.48550/ARXIV.2204.06125}
}

@inproceedings{DBLP:conf/nips/SahariaCSLWDGLA22,
  author       = {Chitwan Saharia and
                  William Chan and
                  Saurabh Saxena and
                  Lala Li and
                  Jay Whang and
                  Emily L. Denton and
                  Seyed Kamyar Seyed Ghasemipour and
                  Raphael Gontijo Lopes and
                  Burcu Karagol Ayan and
                  Tim Salimans and
                  Jonathan Ho and
                  David J. Fleet and
                  Mohammad Norouzi},
  title        = {Photorealistic Text-to-Image Diffusion Models with Deep Language Understanding},
  booktitle    = {NeurIPS},
  year         = {2022},
  url          = {http://papers.nips.cc/paper\_files/paper/2022/hash/ec795aeadae0b7d230fa35cbaf04c041-Abstract-Conference.html}
}

@inproceedings{DBLP:conf/iccv/GandikotaMFB23,
  author       = {Rohit Gandikota and
                  Joanna Materzynska and
                  Jaden Fiotto{-}Kaufman and
                  David Bau},
  title        = {Erasing Concepts from Diffusion Models},
  booktitle    = {ICCV},
  pages        = {2426--2436},
  year         = {2023},
  doi          = {10.1109/ICCV51070.2023.00230},
}

@inproceedings{DBLP:conf/ccs/LiYD0C0024,
  author       = {Xinfeng Li and
                  Yuchen Yang and
                  Jiangyi Deng and
                  Chen Yan and
                  Yanjiao Chen and
                  Xiaoyu Ji and
                  Wenyuan Xu},
  title        = {SafeGen: Mitigating Sexually Explicit Content Generation in Text-to-Image
                  Models},
  booktitle    = {CCS},
  pages        = {4807--4821},
  year         = {2024},
  doi          = {10.1145/3658644.3670295}
}

@inproceedings{DBLP:conf/cvpr/SchramowskiBDK23,
  author       = {Patrick Schramowski and
                  Manuel Brack and
                  Bj{\"{o}}rn Deiseroth and
                  Kristian Kersting},
  title        = {Safe Latent Diffusion: Mitigating Inappropriate Degeneration in Diffusion
                  Models},
  booktitle    = {CVPR},
  pages        = {22522--22531},
  year         = {2023},
  doi          = {10.1109/CVPR52729.2023.02157}
}

@inproceedings{DBLP:conf/iclr/YoonYPYB25,
  author       = {Jaehong Yoon and
                  Shoubin Yu and
                  Vaidehi Patil and
                  Huaxiu Yao and
                  Mohit Bansal},
  title        = {{SAFREE:} Training-Free and Adaptive Guard for Safe Text-to-Image
                  And Video Generation},
  booktitle    = {ICLR},
  year         = {2025},
  url          = {https://openreview.net/forum?id=hgTFotBRKl}
}

@inproceedings{DBLP:conf/wacv/GandikotaOBMB24,
  author       = {Rohit Gandikota and
                  Hadas Orgad and
                  Yonatan Belinkov and
                  Joanna Materzynska and
                  David Bau},
  title        = {Unified Concept Editing in Diffusion Models},
  booktitle    = {WACV},
  pages        = {5099--5108},
  year         = {2024},
  doi          = {10.1109/WACV57701.2024.00503},
}

@inproceedings{DBLP:conf/nips/ParkKCJU23,
  author       = {Yong{-}Hyun Park and
                  Mingi Kwon and
                  Jaewoong Choi and
                  Junghyo Jo and
                  Youngjung Uh},
  title        = {Understanding the Latent Space of Diffusion Models through the Lens
                  of Riemannian Geometry},
  booktitle    = {NeurIPS},
  year         = {2023},
  url          = {http://papers.nips.cc/paper\_files/paper/2023/hash/4bfcebedf7a2967c410b64670f27f904-Abstract-Conference.html}
}

@inproceedings{DBLP:conf/icml/RadfordKHRGASAM21,
  author       = {Alec Radford and
                  Jong Wook Kim and
                  Chris Hallacy and
                  Aditya Ramesh and
                  Gabriel Goh and
                  Sandhini Agarwal and
                  Girish Sastry and
                  Amanda Askell and
                  Pamela Mishkin and
                  Jack Clark and
                  Gretchen Krueger and
                  Ilya Sutskever},
  title        = {Learning Transferable Visual Models From Natural Language Supervision},
  booktitle    = {ICML},
  volume       = {139},
  pages        = {8748--8763},
  year         = {2021},
  url          = {http://proceedings.mlr.press/v139/radford21a.html}
}

@article{DBLP:journals/corr/abs-2210-04610,
  author       = {Javier Rando and
                  Daniel Paleka and
                  David Lindner and
                  Lennart Heim and
                  Florian Tram{\`{e}}r},
  title        = {Red-Teaming the Stable Diffusion Safety Filter},
  journal      = {CoRR},
  volume       = {abs/2210.04610},
  year         = {2022},
  doi          = {10.48550/ARXIV.2210.04610}
}

@article{liu2026saferedir,
  title={SafeRedir: Prompt Embedding Redirection for Robust Unlearning in Image Generation Models},
  author={Liu, Renyang and Chen, Kangjie and Qiu, Han and Zhang, Jie and Lam, Kwok-Yan and Zhang, Tianwei and Ng, See-Kiong},
  journal={arXiv preprint arXiv:2601.08623},
  year={2026}
}

@inproceedings{DBLP:conf/nips/HoJA20,
  author       = {Jonathan Ho and
                  Ajay Jain and
                  Pieter Abbeel},
  title        = {Denoising Diffusion Probabilistic Models},
  booktitle    = {NeurIPS},
  year         = {2020},
  url          = {https://proceedings.neurips.cc/paper/2020/hash/4c5bcfec8584af0d967f1ab10179ca4b-Abstract.html}
}

@inproceedings{DBLP:conf/iclr/SongME21,
  author       = {Jiaming Song and
                  Chenlin Meng and
                  Stefano Ermon},
  title        = {Denoising Diffusion Implicit Models},
  booktitle    = {ICLR},
  year         = {2021},
  url          = {https://openreview.net/forum?id=St1giarCHLP}
}

@article{DBLP:journals/corr/abs-2207-12598,
  author       = {Jonathan Ho and
                  Tim Salimans},
  title        = {Classifier-Free Diffusion Guidance},
  journal      = {CoRR},
  volume       = {abs/2207.12598},
  year         = {2022},
  doi          = {10.48550/ARXIV.2207.12598}
}

@inproceedings{DBLP:conf/nips/GoodfellowPMXWOCB14,
  author       = {Ian J. Goodfellow and
                  Jean Pouget{-}Abadie and
                  Mehdi Mirza and
                  Bing Xu and
                  David Warde{-}Farley and
                  Sherjil Ozair and
                  Aaron C. Courville and
                  Yoshua Bengio},
  title        = {Generative Adversarial Nets},
  booktitle    = {NeurIPS},
  pages        = {2672--2680},
  year         = {2014},
  url          = {https://proceedings.neurips.cc/paper/2014/hash/5ca3e9b122f61f8f06494c97b1afccf3-Abstract.html},
}

@article{Kingma2013AutoEncodingVB,
  title={Auto-Encoding Variational Bayes},
  author={Diederik P. Kingma and Max Welling},
  journal={CoRR},
  year={2013},
  volume={abs/1312.6114},
  url={https://api.semanticscholar.org/CorpusID:216078090}
}

@INPROCEEDINGS{10646735,
  author={Yang, Yuchen and Hui, Bo and Yuan, Haolin and Gong, Neil and Cao, Yinzhi},
  booktitle={2024 IEEE Symposium on Security and Privacy (SP)}, 
  title={SneakyPrompt: Jailbreaking Text-to-image Generative Models}, 
  year={2024},
  pages={897-912},
  doi={10.1109/SP54263.2024.00123}
}

@article{Tsai2023RingABellHR,
  title={Ring-A-Bell! How Reliable are Concept Removal Methods for Diffusion Models?},
  author={Yu-Lin Tsai and Chia-Yi Hsu and Chulin Xie and Chih-Hsun Lin and Jia-You Chen and Bo Li and Pin-Yu Chen and Chia-Mu Yu and Chun-ying Huang},
  journal={ArXiv},
  year={2023},
  volume={abs/2310.10012},
  url={https://api.semanticscholar.org/CorpusID:264146485}
}

@inproceedings{Chin2023Prompting4DebuggingRT,
  title={Prompting4Debugging: Red-Teaming Text-to-Image Diffusion Models by Finding Problematic Prompts},
  author={Zhi-Yi Chin and Chieh-Ming Jiang and Ching-Chun Huang and Pin-Yu Chen and Wei-Chen Chiu},
  booktitle={ICML},
  year={2023},
  url={https://api.semanticscholar.org/CorpusID:261696559}
}

@article{Yang2023MMADiffusionMA,
  title={MMA-Diffusion: MultiModal Attack on Diffusion Models},
  author={Yijun Yang and Ruiyuan Gao and Xiaosen Wang and Nan Xu and Qiang Xu},
  journal={CVPR},
  year={2023},
  pages={7737-7746},
  url={https://api.semanticscholar.org/CorpusID:265498727}
}

@online{dolphin,
  title = {Dolphin3.0-Llama3.1-8B},
  author = {dphn},
  year = {2025},
  url = {https://huggingface.co/dphn/Dolphin3.0-Llama3.1-8B},
}

@inproceedings{Gaintseva2025CASteerCS,
  title={CASteer: Cross-Attention Steering for Controllable Concept Erasure},
  author={Tatiana Gaintseva and Andreea-Maria Oncescu and Chengcheng Ma and Ziquan Liu and Martin Benning and Gregory G. Slabaugh and Jiankang Deng and Ismail Elezi},
  booktitle={ICLR},
  year={2026},
  url={https://api.semanticscholar.org/CorpusID:276961220}
}

@article{Cortes1995SupportVectorN,
  title={Support-Vector Networks},
  author={Corinna Cortes and Vladimir Naumovich Vapnik},
  journal={Machine Learning},
  year={1995},
  volume={20},
  pages={273-297},
  url={https://api.semanticscholar.org/CorpusID:52874011}
}

@article{xiong2025semantic,
  title={Semantic Surgery: Zero-Shot Concept Erasure in Diffusion Models},
  author={Xiong, Lexiang and Liu, Chengyu and Ye, Jingwen and Liu, Yan and Xu, Yuecong},
  booktitle = {CVPR},
  journal={arXiv preprint arXiv:2510.22851},
  year={2025}
}

@article{vaswani2017attention,
  title={Attention is all you need},
  author={Vaswani, Ashish and Shazeer, Noam and Parmar, Niki and Uszkoreit, Jakob and Jones, Llion and Gomez, Aidan N and Kaiser, {\L}ukasz and Polosukhin, Illia},
  journal={Advances in neural information processing systems},
  volume={30},
  year={2017}
}

@article{raffel2020exploring,
  title={Exploring the limits of transfer learning with a unified text-to-text transformer},
  author={Raffel, Colin and Shazeer, Noam and Roberts, Adam and Lee, Katherine and Narang, Sharan and Matena, Michael and Zhou, Yanqi and Li, Wei and Liu, Peter J},
  journal={Journal of machine learning research},
  volume={21},
  number={140},
  pages={1--67},
  year={2020}
}

@online{stable-diffusion-v1.5,
    title={Stable-Diffusion-V1-5},
    author={stable-diffusion-v1-5},
    year={2022},
    url={https://huggingface.co/stable-diffusion-v1-5/stable-diffusion-v1-5},
}

@online{i2p,
    title={I2P},
    author={AIML-TUDA},
    year={2023},
    url={https://huggingface.co/datasets/AIML-TUDA/i2p},
}

@online{sneakyprompt,
    title={SneakyPrompt},
    author={Yuchen413},
    year={2024},
    url={hhttps://github.com/Yuchen413/text2image_safety},
}

@online{rab,
    title={Ring-A-Bell},
    author={chiayi-hsu},
    year={2024},
    url={https://github.com/chiayi-hsu/Ring-A-Bell},
}

@online{mma,
    title={MMA-Diffusion},
    author={YijunYang280},
    year={2024},
    url={https://huggingface.co/datasets/YijunYang280/MMA-Diffusion-NSFW-adv-prompts-benchmark},
}

@online{p4d,
    title={P4D},
    author={joycenerd},
    year={2024},
    url={https://huggingface.co/datasets/joycenerd/p4d},
}

@online{coco,
    title={COCO-Karpathy},
    author={yerevann},
    year={2022},
    url={https://huggingface.co/datasets/yerevann/coco-karpathy},
}

@article{Zhang2023GPT4VisionAA,
  title={GPT-4V(ision) as a Generalist Evaluator for Vision-Language Tasks},
  author={Xinlu Zhang and Yujie Lu and Weizhi Wang and An Yan and Jun Yan and Li Qin and Heng Wang and Xifeng Yan and William Yang Wang and Linda Ruth Petzold},
  journal={ArXiv},
  year={2023},
  volume={abs/2311.01361},
  url={https://api.semanticscholar.org/CorpusID:264935635}
}

@InProceedings{Wang_2025_CVPR,
    author    = {Wang, Zhenting and Hu, Shuming and Zhao, Shiyu and Lin, Xiaowen and Juefei-Xu, Felix and Li, Zhuowei and Han, Ligong and Subramanyam, Harihar and Chen, Li and Chen, Jianfa and Jiang, Nan and Lyu, Lingjuan and Ma, Shiqing and Metaxas, Dimitris N. and Jain, Ankit},
    title     = {MLLM-as-a-Judge for Image Safety without Human Labeling},
    booktitle = {CVPR},
    month     = {June},
    year      = {2025},
    pages     = {14657-14666}
}

@online{qwen,
    title={Qwen2.5-VL-7B-Instruct},
    author={Qwen},
    year={2025},
    url={https://huggingface.co/Qwen/Qwen2.5-VL-7B-Instruct},
}

@InProceedings{10.1007/978-3-319-10602-1_48,
author="Lin, Tsung-Yi
and Maire, Michael
and Belongie, Serge
and Hays, James
and Perona, Pietro
and Ramanan, Deva
and Doll{\'a}r, Piotr
and Zitnick, C. Lawrence",
title="Microsoft COCO: Common Objects in Context",
booktitle="Computer Vision -- ECCV 2014",
year="2014",
publisher="Springer International Publishing",
pages="740--755",
isbn="978-3-319-10602-1"
}

@article{DBLP:journals/corr/HeuselRUNKH17,
  author       = {Martin Heusel and
                  Hubert Ramsauer and
                  Thomas Unterthiner and
                  Bernhard Nessler and
                  G{\"{u}}nter Klambauer and
                  Sepp Hochreiter},
  title        = {GANs Trained by a Two Time-Scale Update Rule Converge to a Nash Equilibrium},
  year         = {2017},
  url          = {http://arxiv.org/abs/1706.08500},
  booktitle    = {NeurIPS}
}

@article{DBLP:journals/corr/abs-1801-03924,
  author       = {Richard Zhang and
                  Phillip Isola and
                  Alexei A. Efros and
                  Eli Shechtman and
                  Oliver Wang},
  title        = {The Unreasonable Effectiveness of Deep Features as a Perceptual Metric},
  year         = {2018},
  url          = {http://arxiv.org/abs/1801.03924},
  booktitle    = {CVPR}
}

@online{nudenet,
    title={NudeNet},
    author={notAI-tech},
    year={2024},
    url={https://github.com/notAI-tech/NudeNet},
}

@article{szegedy2013intriguing,
  title={Intriguing properties of neural networks},
  author={Szegedy, Christian and Zaremba, Wojciech and Sutskever, Ilya and Bruna, Joan and Erhan, Dumitru and Goodfellow, Ian and Fergus, Rob},
  journal={arXiv preprint arXiv:1312.6199},
  year={2013}
}

@article{liu2025token,
  title={Token-level constraint boundary search for jailbreaking text-to-image models},
  author={Liu, Jiangtao and Wang, Zhaoxin and Wang, Handing and Tian, Cong and Jin, Yaochu},
  journal={arXiv preprint arXiv:2504.11106},
  year={2025}
}

@article{wang2026safeneuron,
  title={SafeNeuron: Neuron-Level Safety Alignment for Large Language Models},
  author={Wang, Zhaoxin and Liang, Jiaming and Zhu, Fengbin and Zhao, Weixiang and Fang, Junfeng and Ji, Jiayi and Wang, Handing and Chua, Tat-Seng},
  journal={arXiv preprint arXiv:2602.12158},
  year={2026}
}

@inproceedings{wang2024preventing,
  title={Preventing catastrophic overfitting in fast adversarial training: A bi-level optimization perspective},
  author={Wang, Zhaoxin and Wang, Handing and Tian, Cong and Jin, Yaochu},
  booktitle={European Conference on Computer Vision},
  pages={144--160},
  year={2024},
  organization={Springer}
}

@article{zou2023universal,
  title={Universal and transferable adversarial attacks on aligned language models},
  author={Zou, Andy and Wang, Zifan and Carlini, Nicholas and Nasr, Milad and Kolter, J Zico and Fredrikson, Matt},
  journal={arXiv preprint arXiv:2307.15043},
  year={2023}
}

\end{document}